\documentclass[runningheads]{llncs}
\usepackage{graphicx}
\usepackage[dvipsnames]{xcolor}
\usepackage{bm}
\usepackage{mathtools}
\usepackage{multirow}

\begin{document}
\title{Single-Image Super-Resolution Reconstruction based on the Differences of Neighboring Pixels\thanks{Supported by organization x.}}
\titlerunning{SISR Reconstruction based on the Differences of Neighboring Pixels}
%
%
\author{Huipeng Zheng\inst{1} \and Lukman Hakim\inst{1} \and
Takio Kurita\inst{2} \and Junichi Miyao\inst{2}}
\authorrunning{Zheng et al.}
%
\institute{Department of Information Engineering, 
Hiroshima University, 1-4-1 Kagamiyama, Higashi-Hiroshima, 739-8527 Japan \and Graduate School of Advanced Science and Engineering, Hiroshima University, 1-4-1 Kagamiyama, Higashi-Hiroshima-shi, Hiroshima739-8527, Japan
\email{\{m191504,lukman-hakim,tkurita, miyao\}@hiroshima-u.ac.jp}}
\maketitle              
\begin{abstract}
The deep learning technique was used to increase the performance of single image super-resolution (SISR). However, most existing CNN-based SISR approaches primarily focus on establishing deeper or larger networks to extract more significant high-level features.
Usually, the pixel-level loss between the target high-resolution image and the estimated image is used, but the neighbor relations between pixels in the image are seldom used. 
On the other hand, according to observations, a pixel's neighbor relationship contains rich information about the spatial structure, local context, and structural knowledge.
Based on this fact, in this paper, we utilize pixel's neighbor relationships in a different perspective, and we propose the differences of neighboring pixels to regularize the CNN by constructing a graph from the estimated image and the ground-truth image. The proposed method outperforms the state-of-the-art methods in terms of quantitative and qualitative evaluation of the benchmark datasets.

\end{abstract}
\section{Introduction}
Single-Image Super-Resolution (SISR) is a technique to reconstruct a high-resolution (HR) image from a low-resolution (LR) image. 
The challenges problem in the super-resolution task is the ill-pose problem. Many SISR techniques have been developed to address this challenge, including interpolation-based\cite{zhou2012,anbarjafari2010}, reconstruction-based\cite{zhang2012}, and deep learning-based methods\cite{dong2014}.

Even though CNN-based SISR has significantly improved learning-based approaches bringing good performances, existing SR models based on CNN still have several drawbacks. Most SISR techniques based on CNN are primarily concerned with constructing deeper or larger networks to acquire more meaningful high-level features.
Usually, we use the pixel-level loss between the target high-resolution image and the estimated image and neglect the neighbor relations between pixels. 

Basically, natural images have a strong pixel neighbor relationship. It means that a pixel has a strong correlation with its neighbors, but a low correlation with or is largely independent of pixels further away\cite{zhang2018}. In addition, the neighboring relationship of a pixel also contains rich information about the spatial structure, local context, and structural knowledge\cite{zhou2020}. Based on this fact, the authors proposed to introduce the pixel neighbor relationships as a regularizer in the loss function of CNN and applied 
for Anime-like Images Super-Resolution and Fundus Image Segmentation\cite{hakim2021}. 
The regularizer is named Graph Laplacian Regularization based on the Differences of Neighboring Pixels (GLRDN).
The GLRDN is essentially deriving from the graph theory approach. The graph is constructed from the estimated image and the ground-truth image. 
The graphs use the pixel as a node and the edge represented by the "differences" of a neighboring pixel. 
The basic idea is that the differences between the neighboring pixels in the estimated images should be close to the differences in the ground-truth image. 


This study propose the GLRDN for general single image super-resolution and show the effectiveness of the proposed approach by introducing the GLRDN to the state-of-the-art SISR methods (EDSR\cite{lim2017} and RCAN\cite{zhang2018rcan}). 
The proposed GLRDN can combine with the existing CNN-based SISR methods as a regularizer by simply adding the GLRDN term into their loss functions.
We can easily improve the quality of the estimated super-resolution image of the existing SISR methods.

The contribution of this paper can summarize as follow : (1) Proposed GLRDN to capture the relationship between neighboring pixels for general single image super-resolution; (2) Analyzed the baseline architecture with and without our regularizer; (3) Explored our proposed methods with  state-of-the-art methods in single image super-resolution.

The structure of this paper is as follows. In Section 2, we presented some related methods with our work. In section 3, we explain the proposed method. The results and experiments are detailed in Section 4. Finally, section 5 is presented the conclusion of this study.

\section{Related Work}

\subsection{Graph Laplacian Regularization based on the Differences of Neighboring Pixels}
The GLRDN was proposed by Hakim et al.\cite{hakim2021}. This regularizer uses the graph theory approach to capture the relationship of the difference between pixels. Assume that we have two images, estimated image $y$ and target image $t$. Then $G=(V,E)$ constructed be a graph where $V=\{i|i=1,\ldots, N\}$ is the set of the pixel indices with $N$ pixels and the $E=\{(i,j)| i,j \in V \}$ is the neighboring relations between the pixels. Furthermore,  the differences of neighboring pixels of two images $s_G$ are given as 

\begin{align}
S_G(\bm{t}, \bm{y}) &= \sum_{(i,j) \in E} \{(t_{i} - t_{j}) - (y_{i} - y_{j})\}^2 \nonumber \\
&= \sum_{(i,j) \in E} (\Delta t_{ij} - \Delta y_{ij})^2  \nonumber \\
&= (\Delta \bm{t} - \Delta \bm{y})^T (\Delta \bm{t} - \Delta \bm{y}) \nonumber \\
&= (B \bm{t} - B \bm{y})^T (B \bm{t} - B \bm{y}) \nonumber \\
&= (\bm{t} - \bm{y})^T B^T B (\bm{t} - \bm{y}) \nonumber \\
&= (\bm{t} - \bm{y})^T L (\bm{t} - \bm{y})
\end{align}
where $B$ is incident matrix and $L$ is the Laplacian matrix that is defined from the identity matrix.


\section{Method}
This study aims to capture neighboring pixel's relationships from the reconstructed image estimated from the LR image and the HR images and minimize the differences of the adjacent pixels differences.
As a result, the loss is defined as the squared errors of the differencess between the predicted image and HR images.
In the following sections, we will go through the specifics of the proposed approach.

\subsection{Estimation of the Differences Neighboring Pixels}

Let us consider the set of training samples $X = \{(\bm{x}_m,\bm{t}_m)|m=1,...,M\}$ where $\bm{x}_m$ is a $m^{th}$ input image and $\bm{t}_m$ is the $m^{th}$ target image. $M$ is define as the total of images in training samples. The network is trained to predict the output HR image $\bm{y}_m$ from the $m^{th}$ input LR image $\bm{x}_m$. 

The GLRDN is defined as the graph, which is pixels as nodes and sum of the squared differences of the differences of neighboring pixels between the target image $\bm{t}_m$ and the estimated images $\bm{y}_m$ as edges.   
Then the GLRDN is given as
\begin{align}
S_{G} &= \sum_{m=1}^M S_G(\bm{t}_m, \bm{y}_m) 
= \sum^M_{m=1}(\bm{t}_m-\bm{y}_m)^T L(\bm{t}_m-\bm{y}_m)
\end{align}

This measure $S_{G}$ becomes small if the neighboring relations of the pixels in the estimated output images are similar to those of the target images.

\subsection{CNN-based Super-Resolution with GLRDN}

\begin{figure}[t!]
\includegraphics[width=\textwidth]{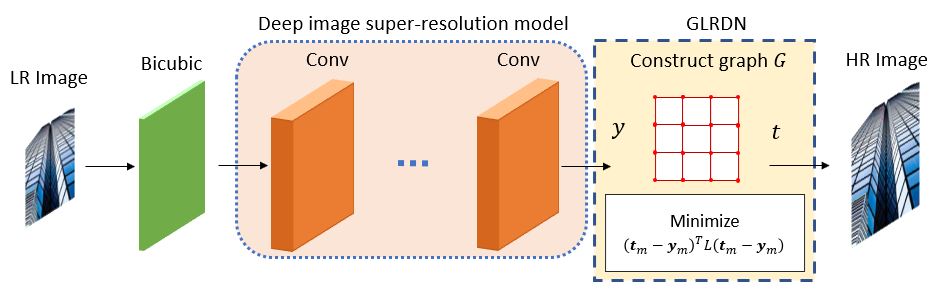}
\caption{Illustration of the proposed method on CNN-based Super-Resolution} \label{architecture}
\end{figure}

We can apply the proposed GLRDN to any existing CNN-based Super-Resolution algorithms by simply adding the GLRDN term in the loss function for the training. The proposed method is illustrated in Fig. \ref{architecture}.
The CNN-based Super-Resolution is trained to estimate the HR image as $y$ for a given LR input image $x$. The first convolutional layer retrieves a series of feature maps. The second layer non-linearly transfers these feature maps to high-resolution patch representations. To construct the final high-resolution image, the last layer integrates the estimates within a spatial neighborhood. 

In the Super-resolution task, using the Sum Squared Error (SSE) as the objective function is common. The Sum Squared Error is given by 
\begin{equation}
    E_{sse} = \sum^M_{m=1} (\bm{t}_m-\bm{y}_m)^2
\end{equation}

For the training of the parameters of the network, we combine the SSE loss with the regularization term as
\begin{equation}
Q_{sr}  = E_{sse}+\lambda S_{G}
\label{eq_loss}
\end{equation}
where $\lambda$ is a parameter to adjust the regularization. The network learning process is more robust by adding the term regularization because it considers the relationship between pixels rather than just comparing pixels with pixels.

\begin{figure}[t!]
\includegraphics[width=\textwidth]{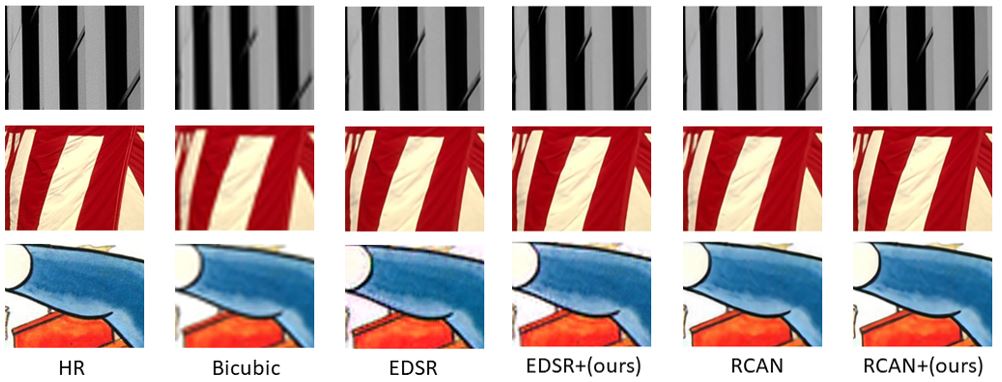}
\caption{Visual comparison of our proposed methods on Urban100, B100, and Manga109 datasets.} \label{results-sota}
\end{figure}

\section{Experiments}



\subsection{Experimental Setting}

We adopt the EDSR and RCAN as our baseline models due to their great performance on image super-resolution tasks. In all these settings, we compare the performance with and without our regularizer. We set 300 epochs and batch size to 16. We set the learning rate to $10^{-4}$ and divided at every $2\times10^{5}$ minibatch.

Our experiments are performed under the $\times2$, $\times3$, $\times4$ scale factor. During training, we use the RGB input patches with the size of $48\times48$ in each batch. Augmentation technique also used on the training images by rotating $90^{\circ}$, $180^{\circ}$, $270^{\circ}$, and flipped randomly. This experiments implemented on DIV2K\cite{agustsson2017}, Set5\cite{set5}, Set14\cite{set14}, B100\cite{martin2001}, Urban100\cite{huang2015}, and Manga109\cite{manga109} datasets. We asses the improvement of our method using PSNR and SSIM measurements.

\begin{table}[!t]
\centering
\caption{Ablation study on Set5, Set14, and B100 datasets.}
\label{tab:ablation1}
\begin{tabular}{|p{2.2cm}|p{0.8cm}|p{1.2cm}|p{1.2cm}|p{1.2cm}|p{1.2cm}|p{1.2cm}|p{1.2cm}|}
\hline
\multirow{2}{*}{Method} & \multirow{2}{*}{$\lambda$} & \multicolumn{2}{c|}{Set5}        & \multicolumn{2}{c|}{Set14}       & \multicolumn{2}{c|}{B100} \\ \cline{3-8} 
          &     & PSNR  & SSIM   & PSNR  & SSIM   & PSNR  & SSIM   \\ \hline
Bicubic   & -   & 28.42 & 0.8104 & 26.00 & 0.7027 & 25.96 & 0.6675 \\
EDSR      & 0   & 30.89 & 0.8683 & 27.66 & 0.7515 & 27.12 & 0.7159 \\
EDSR+ours & 0.1 & 31.69 & 0.8851 & 28.15 & 0.7655 & 27.49 & 0.7279 \\
EDSR+ours               & 1                          & \textbf{31.75} & \textbf{0.8863} & \textbf{28.19} & \textbf{0.7663} & \textbf{27.52}  & \textbf{0.7643}  \\
EDSR+ours & 5   & 31.74 & 0.8857 & 28.18 & 0.7653 & \textbf{27.52} & 0.7262 \\
EDSR+ours & 10  & 31.75 & 0.8855 & 28.18 & 0.7641 & \textbf{27.52} & 0.7252 \\
EDSR+ours & 100 & 31.65 & 0.8840 & 28.12 & 0.7620 & 27.49 & 0.7230 \\ \hline
\end{tabular}
\end{table}

\begin{table}[!t]
\centering
\caption{Ablation study on Urban100 and Manga109 datasets.}
\label{tab:ablation2}

\begin{tabular}{|p{2.2cm}|p{0.8cm}|p{1.2cm}|p{1.2cm}|p{1.2cm}|p{1.2cm}|}
\hline
\multirow{2}{*}{Method} & \multirow{2}{*}{$\lambda$} & \multicolumn{2}{c|}{Urban100} & \multicolumn{2}{c|}{Manga109} \\ \cline{3-6} 
          &     & PSNR  & SSIM            & PSNR           & SSIM            \\ \hline
Bicubic   & -   & 23.14 & 0.6577          & 24.89          & 0.7866          \\
EDSR      & 0   & 25.12 & 0.7445          & 29.68          & 0.8999          \\
EDSR+ours & 0.1 & 25.83 & 0.7749          & 30.84          & 0.9061          \\
EDSR+ours & 1   & 25.92 & 0.7749          & \textbf{30.95} & \textbf{0.9084} \\
EDSR+ours & 5   & \textbf{25.95} & \textbf{0.7767} & 30.91          & 0.9072          \\
EDSR+ours & 10  & \textbf{25.95} & 0.7762          & 30.86          & 0.9062          \\
EDSR+ours & 100 & 25.92 & 0.7736          & 30.84          & 0.9042          \\ \hline
\end{tabular}
\end{table}




\section{Result and Discussion}

\begin{table}[!t]
\caption{Performance of our proposed method compared with the state of the art method.}
\label{tab:sr-sota}
\resizebox{\textwidth}{!}{%
\begin{tabular}{|p{2.2cm}|p{0.8cm}|p{1cm}|p{1.1cm}|p{1cm}|p{1.1cm}|p{1cm}|p{1.1cm}|p{1cm}|p{1.1cm}|p{1cm}|p{1.1cm}|}
\hline
\multirow{2}{*}{Method} & \multirow{2}{*}{scale} & \multicolumn{2}{c|}{Set5} & \multicolumn{2}{c|}{Set14} & \multicolumn{2}{c|}{B100} & \multicolumn{2}{c|}{Urban100} & \multicolumn{2}{c|}{Manga109} \\ \cline{3-12} 
                        &                        & PSNR        & SSIM        & PSNR         & SSIM        & PSNR        & SSIM        & PSNR          & SSIM          & PSNR          & SSIM          \\ \hline
\end{tabular}}
\resizebox{\textwidth}{!}{%
\begin{tabular}{|p{2.2cm}|p{0.8cm}|p{1cm}|p{1.1cm}|p{1cm}|p{1.1cm}|p{1cm}|p{1.1cm}|p{1cm}|p{1.1cm}|p{1cm}|p{1.1cm}|}
\hline
Bicubic     & x2 & 33.66 & 0.9299 & 30.24 & 0.8688  & 29.56 & 0.8431 & 26.88 & 0.8403 & 30.80  & 0.9339 \\
SRCNN       & x2 & 36.66 & 0.9542 & 32.45 & 0.9067 & 31.36 & 0.8879 & 29.50 & 0.8946 & 35.60 & 0.9663 \\
FSRCNN      & x2 & 37.05 & 0.9560 & 32.66 & 0.9090 & 31.53 & 0.8920 & 29.88 & 0.9020 & 36.67 & 0.9710 \\
VDSR        & x2 & 37.53 & 0.9590 & 33.05 & 0.9130 & 31.90 & 0.8960 & 30.77 & 0.9140 & 37.22  & 0.9750 \\
LapSRN      & x2 & 37.52 & 0.9591 & 33.08 & 0.9130 & 31.08 & 0.8950 & 30.41 & 0.9101 & 37.27 & 0.9740 \\
MemNet      & x2 & 37.78 & 0.9597 & 33.28 & 0.9142 & 32.08 & 0.8978 & 31.31 & 0.9195 & 37.72 & 0.9740 \\
EDSR        & x2 & 38.07 & 0.9606 & 33.65 & 0.9167 & 32.20 & 0.9004 & 31.88 & 0.9214 & 38.22 & 0.9763 \\
SRMDNF      & x2 & 37.79 & 0.9601 & 33.32 & 0.9159 & 32.05 & 0.8985 & 31.33 & 0.9204 & 38.07 & 0.9761 \\
D-DBPN      & x2 & 38.09 & 0.9600 & 33.85 & 0.9190 & 32.27 & 0.9006 & 32.55 & 0.9324 & 38.89 & 0.9775 \\
RDN         & x2 & 38.24 & 0.9614 & 34.01 & 0.9212 & 32.34 & 0.9017 & 32.89 & 0.9353 & 39.18 & 0.9780 \\
RCAN        & x2 & 38.25 & 0.9608 & 34.08 & 0.9213 & 32.38 & 0.9020 & 33.29 & 0.9363 & 39.22 & 0.9778 \\
EDSR+(ours) & x2 & 38.17 & 0.9610 & 33.74 & 0.9182 & 32.25 & 0.9000 & 31.96 & 0.9248 & 38.57 & 0.9764 \\
RCAN+(ours) & x2 & \textbf{38.31} & \textbf{0.9612} & \textbf{34.20} & \textbf{0.9222} & \textbf{32.39} & \textbf{0.9022} & \textbf{33.30} & \textbf{0.9369} & \textbf{39.27} & \textbf{0.9781} \\ \hline
\end{tabular}}
\resizebox{\textwidth}{!}{%
\begin{tabular}{|p{2.2cm}|p{0.8cm}|p{1cm}|p{1.1cm}|p{1cm}|p{1.1cm}|p{1cm}|p{1.1cm}|p{1cm}|p{1.1cm}|p{1cm}|p{1.1cm}|}
\hline
Bicubic     & x3 & 30.39 & 0.8682 & 27.55  & 0.7742 & 27.21 & 0.7385 & 24.46 & 0.7349 & 26.95 & 0.8556 \\
SRCNN       & x3 & 32.75 & 0.9090 & 29.30 & 0.8215 & 28.41 & 0.7863 & 26.24 & 0.7989 & 30.48 & 0.9117 \\
FSRCNN      & x3 & 33.18 & 0.9140 & 29.37 & 0.8240 & 28.53 & 0.7910 & 26.43 & 0.8080 & 31.10 & 0.9210 \\
VDSR        & x3 & 33.67 & 0.9210 & 29.78 & 0.8320 & 28.83 & 0.7990 & 27.14 & 0.8290 & 32.01 & 0.9340 \\
LapSRN      & x3 & 33.82  & 0.9227 & 29.87 & 0.8320 & 28.82 & 0.7980 & 27.07 & 0.8280 & 32.21 & 0.9350 \\
MemNet      & x3 & 34.09 & 0.9248 & 30.00 & 0.8350 & 28.96 & 0.8001 & 27.56  & 0.8376 & 32.51 & 0.9369 \\
EDSR        & x3 & 34.26 & 0.9252 & 30.08 & 0.8418 & 29.20 & 0.8106 & 28.48 & 0.8638 & 33.20 & 0.9415 \\
SRMDNF      & x3 & 34.12 & 0.9254 & 30.04 & 0.8382 & 28.97 & 0.8025 & 27.57 & 0.8398 & 33.00 & 0.9403 \\
RDN         & x3 & 34.71 & 0.9296 & 30.57 & 0.8468 & 29.26 & 0.8093 & 28.80 & 0.8653 & 34.13 & 0.9484 \\
RCAN        & x3 & 34.79 & 0.9255 & 30.39 & 0.8374 & 29.40 & 0.8158 & 29.24 & 0.8804 & 33.99 & 0.9469 \\
EDSR+(ours) & x3 & 34.41 & 0.9253 & 30.18 & 0.8443 & 29.27 & 0.8141 & 28.49 & 0.8672 & 33.76 & 0.9416 \\
RCAN+(ours) & x3 & \textbf{34.85} & \textbf{0.9259} & \textbf{30.50} & \textbf{0.8392} & \textbf{29.41} & \textbf{0.8186} & \textbf{29.25} & \textbf{0.8838} & \textbf{34.15} & \textbf{0.9484} \\ \hline
\end{tabular}}
\resizebox{\textwidth}{!}{%
\begin{tabular}{|p{2.2cm}|p{0.8cm}|p{1cm}|p{1.1cm}|p{1cm}|p{1.1cm}|p{1cm}|p{1.1cm}|p{1cm}|p{1.1cm}|p{1cm}|p{1.1cm}|}
\hline
Bicubic     & x4 & 28.42 & 0.8104 & 26.00 & 0.7027 & 25.96 & 0.6675 & 23.14 & 0.6577 & 24.89 & 0.7866 \\
SRCNN       & x4 & 30.48 & 0.8628 & 27.50 & 0.7513 & 26.90 & 0.7101 & 24.52 & 0.7221 & 27.58 & 0.8555 \\
FSRCNN      & x4 & 30.72 & 0.8660 & 27.61 & 0.7550 & 26.98 & 0.7150 & 24.62 & 0.7280 & 27.90 & 0.8610 \\
VDSR        & x4 & 31.35 & 0.8830 & 28.02 & 0.7680 & 27.29 & 0.7260 & 25.18 & 0.7540 & 28.83 & 0.8870 \\
LapSRN      & x4 & 31.54 & 0.8850 & 28.19 & 0.7720 & 27.32 & 0.7270 & 25.21 & 0.7560 & 29.09 & 0.8900 \\
MemNet      & x4 & 31.74 & 0.8893  & 28.26 & 0.7723 & 27.40 & 0.7281 & 25.50 & 0.7630 & 29.42 & 0.8942 \\
EDSR        & x4 & 32.04 & 0.8926 & 28.43 & 0.7755 & 27.70 & 0.7351 & 26.45 & 0.7908 & 30.25 & 0.9028 \\
SRMDNF      & x4 & 31.96 & 0.8925 & 28.35 & 0.7787 & 27.49 & 0.7337 & 25.68 & 0.7731 & 30.09 & 0.9024 \\
D-DBPN      & x4 & 32.47 & 0.8980 & 28.82 & 0.7860 & 27.72 & 0.7400 & 26.38 & 0.7946 & 30.91 & 0.9137 \\
RDN         & x4 & 32.47 & 0.8990 & 28.81 & 0.7871 & 27.72 & 0.7419 & 26.61 & 0.8028 & 31.00 & 0.9151 \\
RCAN        & x4 & 32.78 & 0.8988 & 28.68 & 0.7832 & 27.85 & 0.7418 & 27.07 & 0.8121 & 31.02 & 0.9157 \\
EDSR+(ours) & x4 & 32.21 & 0.8934 & 28.51 & 0.7768 & 27.75 & 0.7369 & 26.52 & 0.7937 & 30.53 & 0.9057 \\
RCAN+(ours) & x4 & \textbf{32.90} & \textbf{0.8992} & \textbf{28.79} & \textbf{0.7849} & \textbf{27.86} & \textbf{0.7423} & \textbf{27.13} & \textbf{0.8139} & \textbf{31.10} & \textbf{0.9163} \\ \hline
\end{tabular}}
\end{table}

\textbf{Ablation Study. }In this part, the ablation study presented the effect of the proposed regularizer. We combined EDSR with our regularizer by setting different $\lambda$. We started with a simple EDSR model by setting the number of layers $B$ = 12 and the number of feature channels $F$ = 64 with a scaling factor of 1. We compared the PSNR/SSIM result on the different testing datasets by setting the scale factor as 4. Table  \ref{tab:ablation1} showing the ablation study on Set5, Set14, and B100 datasets, and Table \ref{tab:ablation2} showing the ablation study on Urban100 and  Manga109 datasets. The best results are highlighted in bold. As shown in Table \ref{tab:ablation1} and Table  \ref{tab:ablation2}, the best parameter $\lambda$ in Eq. \ref{eq_loss} is 1 which highest PSNR and SSIM on Set5, Set14, B100, and Manga109 datasets. Meanwhile, we found that $\lambda$=5 is the best on Urban100 datasets. We obtained these values by performing parameter experiments in the ranges 0 to 100, $\lambda$=0 means we use only EDSR as a baseline without a regularizer. Along with increasing lambda, the stronger the influence of the relationship between pixels in the learning process. Compared to baseline, our approach achieved an improvement of PSNR and SSIM scores over all datasets.
\\

\noindent
\textbf{Comparation with state-of-the-art. }
To know the advantages of our proposed regularizer, we combine our regularizer with EDSR and RCAN and then compare the result with state-of-the-art CNN-based SR methods. Table \ref{tab:sr-sota} summarizes all of the quantitative data for the various scaling factors. The best results are highlighted in bold. Compared to competing approaches, joining RCAN and our methods achieve the best results across all datasets and scaling factors. The qualitative result of our approach is shown in Fig. \ref{results-sota}. To know the differences in detail, we zoomed in on a portion of the image area. Fig. \ref{results-sota} showing our approach demonstrated more realistic visual results compared to other methods on Urban100, B100, and Manga109 datasets. It means the proposed regularizer succeeds in reconstructing the details of the HR image generate from the LR image compared over baseline methods.

\section{Conclusion}
This paper shows that the differences in pixels neighbor relationships can establish the network more robust on super-resolution tasks. Our method employs the adjacent pixels differences as a regularizer with existing CNN-based SISR methods to ensure that the differences between pixels in the estimated image are close to different pixels in the ground truth images. The experimental findings on five datasets demonstrate that our method outperforms the baseline CNN without regularization. Our proposed method generates more detailed visual results and improved PSNR/SSIM scores compared to other state-of-the-art methods. Future work will implement the differences in pixel neighbor relationships as a regularizer on different computer vision tasks.  

\section*{Acknowledgments} 
This work was partly supported by JSPS KAKENHI Grant Number 21K12049.

\end{document}